\documentclass[sn-basic,iicol]{sn-jnl}

\jyear{2021}%

\theoremstyle{thmstyleone}%
\graphicspath{{./figures/}}

\definecolor{revised_color}{HTML}{FAA21A}

\theoremstyle{thmstyletwo}%

\theoremstyle{thmstylethree}%

\begin{document}

\title[Article Title]{A New Dataset and a Distractor-Aware Architecture for Transparent Object Tracking}

\author*[1]{\fnm{Alan} \sur{Lukežič}}\email{alan.lukezic@fri.uni-lj.si}

\author[1]{\fnm{Žiga} \sur{Trojer}}\email{ziga.trojer20@gmail.com}

\author[2]{\fnm{Jiří} \sur{Matas}}\email{matas@fel.cvut.cz}

\author[1]{\fnm{Matej} \sur{Kristan}}\email{matej.kristan@fri.uni-lj.si}

\affil[1]{\orgdiv{Faculty of computer and information science}, \orgname{University of Ljubljana}, \orgaddress{\street{Večna pot 113}, \city{Ljubljana}, \postcode{1000}, \country{Slovenia}}}
 
\affil[2]{\orgdiv{Center for Machine Perception}, \orgname{Czech Technical University in Prague}, \orgaddress{\street{Karlovo namesti 13}, \city{Prague}, \postcode{12135}, \country{Czech Republic}}}

%%==================================%%
%% Abstract %%
%%==================================%%

\abstract{
Performance of modern trackers degrades substantially on transparent objects compared to opaque objects.
This is largely due to two distinct reasons. 
Transparent objects are unique in that their appearance is directly affected by the background. 
Furthermore, transparent object scenes often contain many visually similar objects (distractors), which often lead to tracking failure. 
However, development of modern tracking architectures requires large training sets, which do not exist in transparent object tracking. 
We present two contributions addressing the aforementioned issues. 
We propose the first transparent object tracking \textit{training dataset} Trans2k that consists of over 2k sequences with 104,343 images overall, annotated by bounding boxes and segmentation masks. Standard trackers trained on this dataset consistently improve by up to 16\%. 
Our second contribution is a new distractor-aware transparent object tracker (DiTra) that treats localization accuracy and target identification as separate tasks and implements them by a novel architecture. DiTra sets a new state-of-the-art in transparent object tracking and generalizes well to opaque objects. 
}


\maketitle

%%==================================%%
%% Introduction %%
%%==================================%%
\section{Introduction}  \label{sec:intro}

Visual object tracking is a fundamental computer vision problem with far-reaching applications in human-computer interaction, surveillance, autonomous robotics, and video editing, among others. 
The significant progress observed over the past decade can be attributed to the emergence of challenging evaluation datasets~\citep{otb_pami2015,kristan_vot_tpami2016,got10k,lasot_cvpr19} and diverse training sets~\citep{muller_trackingnet,imagenet_ijcv_2015,got10k}, which have facilitated the end-to-end learning of modern deep tracking architectures.

\begin{figure*}[t]
\centering
\includegraphics[width=\linewidth]{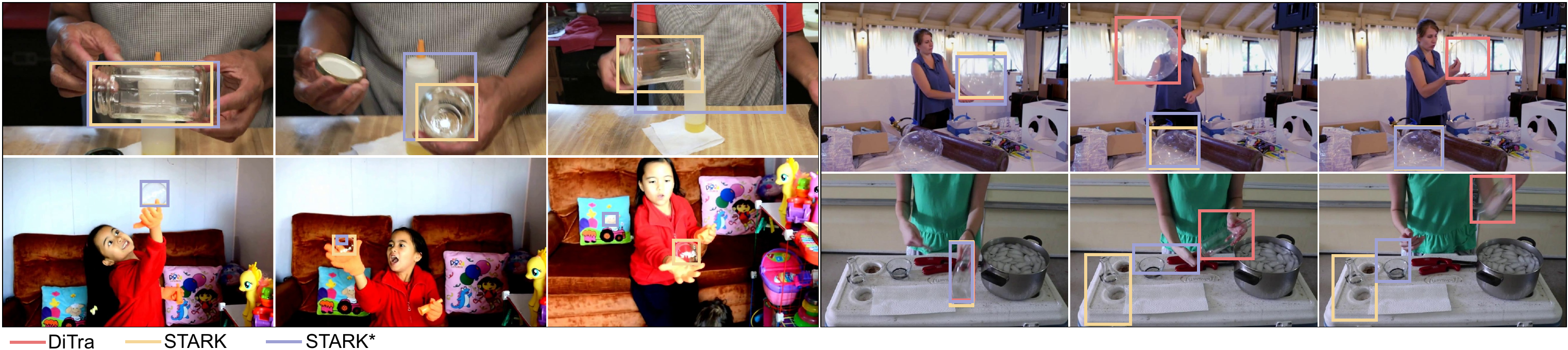}
\caption{
A tracker STARK* trained on opaque objects fails on transparent objects, while its performance remarkably improves after training on the proposed Trans2k dataset (first and second row). Both versions, however, fail in presence of visual distractors (third and fourth row), while the proposed DiTra comfortably tracks due to the new distractor-aware visual model. 
}
\label{fig:first}
\end{figure*}

While numerous benchmarks have addressed opaque objects, tracking of transparent objects has received comparatively little attention. 
These objects are unique due to their reflective nature and dependence on background texture, which reduces the effectiveness of deep features trained for opaque objects, as shown in Figure~\ref{fig:first}. 
 
Transparent objects, e.g., cups and glasses are common in households. 
Thus a household robot or ambient intelligence systems for scene and activity understanding will rely on accurate tracking and localization of such objects.
Furthermore, transparent object localization is already crucial in modern automated end-of-line quality control systems such as in the glassmaking industry.

The recent transparent object tracking benchmark (TOTB), demonstrated that trackers based on deep learning outperform traditional (non-deep learning) methods on transparent objects, even when trained on opaque objects. 
Furthermore, the benchmark revealed that the performance of the state-of-the-art trackers designed for opaque objects drops when applied to transparent objects.
However, these results were obtained without re-training the trackers on representative training sets, which raises the question of whether the observed performance drop is a consequence of a domain shift, rather than an inherent property of the problem. 
Consequently, there is an urgent need for a high-quality transparent object training video dataset to address this question and potentially unlock the power of deep learning trackers. 

In general, the construction of training datasets presents numerous challenges. 
First, the dataset must be large, diverse, and focused on visual attributes and challenging scenarios specific to transparent objects, that are not covered in opaque tracking datasets. 
Second, the targets should be annotated accurately. 
Given these challenges, various sequence selection and annotation protocols have been developed for related problems~\citep{got10k,kristan_vot2013,totb_iccv21,kristan_vot2014}. 
In 6DoF estimation~\citep{Hodan_2020_CVPR,Hodan_2018_ECCV} and scene parsing~\citep{scene_neurips18,sceneparsing_iccv19,kegan_cvpr19}, image rendering has been employed to circumvent these issues. 
While the realism of rendered opaque objects may be limited, transparent objects are unique in that non-textured transparent materials can be rendered faithfully by modern renderers~\citep{blenderproc}. This allows generating highly realistic sequences with precisely specified visual attributes and pixel-level ground truth, free of subjective annotation bias.
We exploit this property and introduce the first transparent object tracking training dataset, Trans2k, which is our first contribution~\footnote{The data that support the findings of this study are openly available in Github repository at~\url{https://github.com/trojerz/Trans2k}}.

An intriguing aspect of videos featuring transparent objects is the frequent presence of multiple visually similar transparent objects, or distractors (Figure~\ref{fig:first}). 
For instance, kitchen or laboratory scenes often contain several glasses and bottles, while crowded scenes are commonplace in industrial manufacturing lines producing identical object types (i.e., distractors). 
Efficient distractor handling mechanisms are thus essential to achieve robust tracking of transparent objects. 

Our second contribution is a new distractor-aware transparent object tracker (DiTra), which addresses the situations when multiple objects, visually similar to the target (distractors), are present in the scene. 
DiTra treats target localization accuracy (i.e., precise bounding box estimation) and localization robustness (i.e., selecting the correct target among similar objects) as distinct problems (Figure~\ref{fig:first}).
A common backbone encodes the image, while separate branches are utilized to extract features specialized for localization accuracy and robustness. 
These features are then fused into target-specific localization features and regressed into a bounding box.
The proposed tracker is able to track an arbitrary transparent object, regardless of  the object category.

In summary, our contributions include: 
(i) Trans2k, the first training dataset for transparent object tracking that unlocks the power of deep trainable trackers and allows for training bounding box or segmentation trackers, and 
(ii) the accuracy/robustness split architecture with the distractor-aware block for computing robust localization features. 

A variety of trackers representing major modern deep learning approaches is evaluated on TOTB~\citep{totb_iccv21}. After re-training on Trans2k, a consistent performance boost (up to 16\%) is observed across all architectures. 
The proposed DiTra outperforms all re-trained trackers, setting a new state-of-the-art on the TOTB benchmark~\cite{totb_iccv21}, making it a strong baseline for this task. 

We initially presented the Trans2k dataset in a conference paper~\cite{trans2k_bmvc2022}. Here we further explore its performance contributions to existing state-of-the-art trackers. 
We also propose a new tracker DiTra specialized for transparent objects, and demonstrate substantial benefits from training on Trans2k.

%%==================================%%
%% Related work %%
%%==================================%%
\section{Related work}  \label{sec:related_work}

\textbf{Object tracking.} Deep trackers excel across various benchmarks~\citep{kristan_vot2020,kristan_vot2021,got10k,lasot_cvpr19,otb_pami2015,totb_iccv21} compared to their hand-crafted counterparts.
Initially, pre-trained general backbones were used for feature extraction, primarily by the discriminative correlation filter (DCF) trackers~\citep{danelljan_eccv2016_ccot,DanelljanCVPR2017,danelljan_eccv2018_updt,danelljan_iccvw2015,liu_icme2021}, which learned a discriminative localization models online during tracking.
Later, backbone end-to-end training techniques that maximize DCF localization were proposed~\citep{Valmadre_2017_CVPR}. 
Most recently, the DCF optimization has been introduced as part of the deep network. Milestone representatives were proposed in~\citep{atom_cvpr19,danelljan_dimp_iccv19,kys_eccv2020}, which also 
proposed a post-processing network for bounding box refinement that accounted for target aspect changes.
In parallel, siamese trackers have been explored and grown into a major tracker design branch. The seminal work~\citep{siamfc_eccvw2016} trained AlexNet-based network~\citep{alexnet_nips12} such that localization accuracy is maximized simply by correlation between a template and search region in feature space. 
These trackers afford fast processing since no training is required during tracking. Siamese trackers were extended by anchor-based region proposal networks~\citep{siamrpn_cvpr2018,siamrpn_cvpr2019} and recently an anchor-free extension has been proposed~\citep{siamban_cvpr20} with improved localization performance. 
Drawing on advances in object detection~\citep{detr_eccv2020}, transformer-based trackers have recently emerged ~\citep{stark_iccv21,transt_cvpr2021,transf_cvpr2021,tomp_cvpr2022}.
These are the current state-of-the-art, and computationally efficient with remarkable real-time performance~\citep{kristan_vot2021}. 

\textbf{Benchmarks.} The developments in visual object tracking have been facilitated by introduction of benchmarks. The first widely-used benchmark~\citep{otb_pami2015,otb_cvpr2010} proposed a dataset and evaluation protocol that allowed standardised comparsion of trackers.
Later, the VOT initiative explored dataset construction as well as performance evaluation protocols for efficient in-depth analysis~\citep{kristan_vot_tpami2016,kristan_vot2013,kristan_vot2014}. Further improvements were made in the subsequent yearly challenges, e.g.,~\citep{kristan_vot2020,kristan_vot2021}. 
With advent of deep learning, tracking training sets have emerged.
\citep{muller_trackingnet} constructed a huge training set from public video repository and applied a semi-automatic annotation. 
Recently,~\citep{got10k} presented ten thousand annotated video sequences, divided into a large training and a smaller evaluation set. 
Concurrently, a long-term tracking benchmark~\citep{lasot_cvpr19} with fifteen pre-defined categories, containing training and test set was proposed.  
All these benchmarks focus on opaque objects, while recently as transparent object tracking evaluation dataset ~\citep{totb_iccv21} has been proposed. However, training datasets for transparent object tracking have not been proposed.

\textbf{Use of synthesis.} 
Rendering has been previously considered in computer vision to avoid costly manual dataset acquisition. In~\citep{video_games_cvpr2018,playing_iccv2017}, synthetic data was generated by a video game engine, which provided an unlimited amount of annotated training data for various computer vision tasks.
A rendered dataset of urban scenes, Synthia~\citep{synthia_cvpr2016}, was shown to substantially improve the trained deep models for semantic segmentation. 
A similar dataset~\citep{synscapes_2018} was proposed for training and evaluation of scene parsing networks.
A fine-grained vegetation and terrain dataset~\citep{metzger_icpr2021} was recently proposed for training drivable surfaces and natural obstacles detection networks in outdoor scenes.
\citep{synthetic_eccv2018} showed that foreground and background should be treated differently when training segmentation on synthetic images.
The benefits of using mixed real and synthetic 6DoF training data has been recently shown in~\citep{hodan_cvpr2020}.
The major 6DoF object detection challenge~\citep{hodan_bop2020} thus provides a combination of real and synthetic images for training as well as evaluation.  
Synthesis has been used in the UAV123 tracking benchmark~\citep{uav_benchmark_eccv2016} in which eight of the sequences are rendered by a game engine. 
A rendering approach was used in~\citep{cehovin_iccv2017}  to parameterize camera motion for fine-grained tracker performance analysis.
However, using synthetic data for training in visual tracking remains unexplored.
 
\textbf{Transparent object sensing.} Highlighting the difference from opaque counterparts, transparent objects have been explored in computer vision in various tasks. 
Recognition of transparent objects was studied in~\citep{fritz_nips2009,maeno_cvpr2013}, while 3D shape estimation and reconstruction of transparent objects on RGB-D images was proposed in~\citep{klank_icra2011,sajjan_icra2020}. 
Segmentation of transparent objects has been studied in~\citep{xu_iccv2015,kalra_cvpr2020}, while a benchmark was proposed in~\citep{xie_eccv2020}. 
All these works consider single-image tasks and little attention has been dedicated to videos. 
In fact, a transparent object tracking benchmark~\citep{totb_iccv21} has been proposed only recently and reported a performance gap between transparent and opaque object tracking. 
However, due to the lack of a dedicated training dataset, the gap source remains unclear. 

%%==================================%%
%% Trans2k %%
%%==================================%%
\section{Trans2k dataset}  \label{sec:dataset}

\begin{figure*}[t]
\centering
\includegraphics[width=\linewidth]{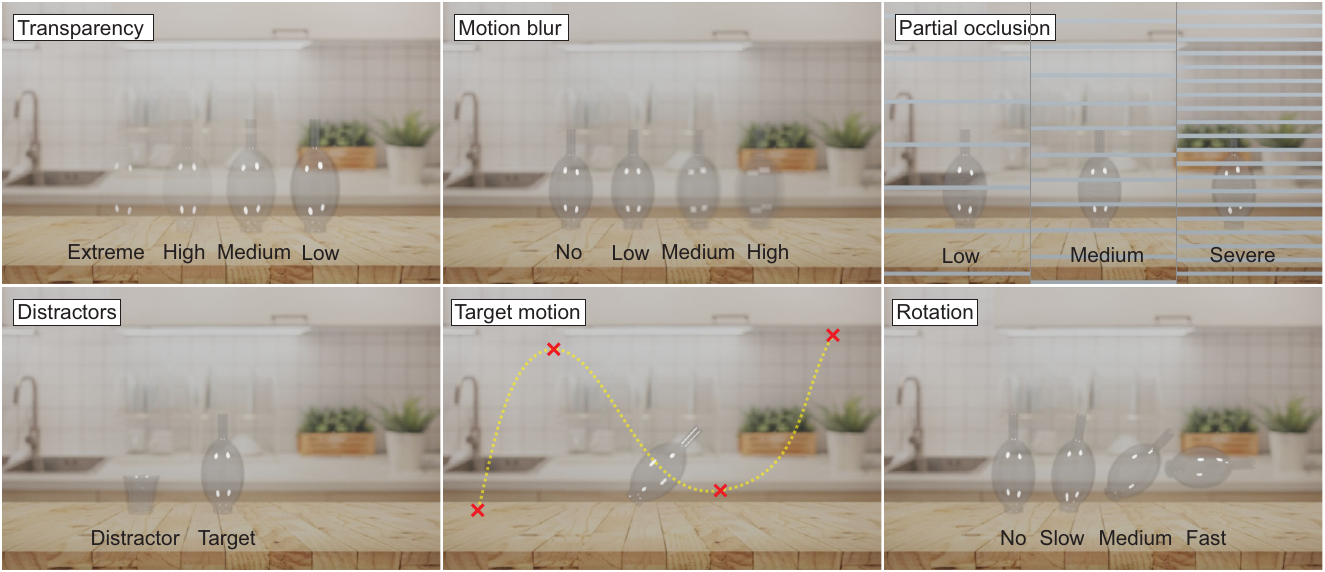}
\caption{Trans2k attribute levels for "Transparency", "Motion blur", "Partial occlusion", "Distractor" (binary), "Target motion" (four control points) and "Rotation".}
\label{fig:parameter_examples}
\end{figure*}

Transparent objects, which are often reflective and glass-like, can be rendered with a high level of realism by the modern photo-realistic rendering engines~\citep{blenderproc}.
In our approach, we first identify and parameterize the sequence attributes specific to transparent objects in Section~\ref{sec:attributes}. 
A BlenderProc-based sequence generator is implemented that enables parameterized sequence rendering. 
Attribute levels useful for learning are identified empirically and the final training dataset is presented in Section~\ref{sec:range_identification}.

\subsection{Parametrization of sequence attributes}  \label{sec:attributes}

An efficient training dataset should reflect the diversity of visual attributes typical for transparent object tracking scenes. After carefully examining various videos of transparent and opaque objects, the following attributes were identified (Figure~\ref{fig:parameter_examples}).  

{\noindent \bf Scene background.} Since background affects the transparent object appearance, a high background diversity is required in training. 
We ensure this by randomly sampling videos from GoT10k~\citep{got10k} training set and use them as backgrounds over which the transparent object is rendered.
Sampled background sequences are highly diverse, including indoor and outdoor environments, and scenes from sports, nature, marine and traffic, to name just a few.

\begin{figure*}[t]
\centering
\includegraphics[width=\linewidth]{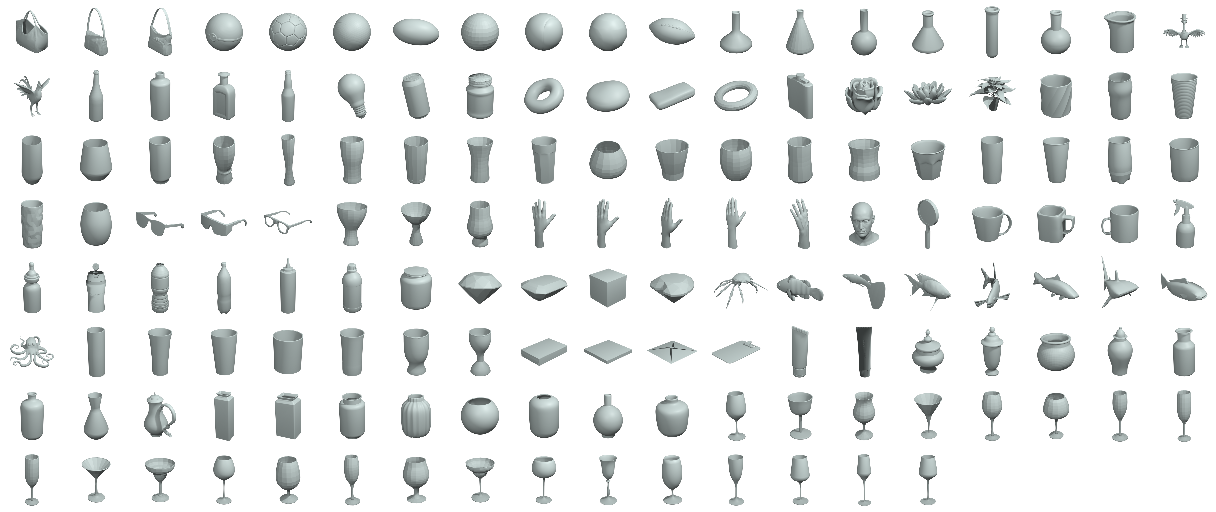}
\caption{A diverse set of object instances used in rendering Trans2k sequences.}
\label{fig:trans2k_objects}
\end{figure*}

{\noindent \bf Object types.} 3D models of 25 object types from open source online repositories are selected with several instances of the same type. 
The set was chosen such to cover a range of nontrivial as well as smooth shapes, with some objects rendered with empty and some with full volume. 
This amounts to 148 object instances, which are visualized in Figure~\ref{fig:trans2k_objects}. 

{\noindent \bf Target motion.} To increase the object-background appearance diversity, the objects are moving in the videos. 
The motion trajectory is generated by a cubic Hermite spline spanned by four uniformly sampled 2D points. 
The motion dynamics is not critical in training, since deep models are typically trained on pairs of image patches cropped at target position. 
Thus a constant velocity is applied.

{\noindent \bf Distractors.} In realistic environments, the target may be surrounded by other visually similar transparent objects (e.g., glasses on a table), which act as distractors. We thus render an additional transparent object following the target object. 
The distractor object is of a different type to keep the appearance-based localization learning task feasible.

{\noindent \bf Transparency.} The transparency level crucially affects the target appearance. We thus identify four levels ranging from clearly visible to nearly invisible.

{\noindent \bf Motion blur.} Fast motions, depending on the aperture speed, result in various levels of blurring. We identify four levels of blur intensity, ranging from no blurring to extreme blurriness.

{\noindent \bf Partial occlusion.} Objects are commonly occluded by other objects in practical situations (e.g., handling of the target). 
We thus simulate partial occlusions by rendering coloured stripe pattern moving across the video frame. 
The stripe width is fixed, while the occlusion intensity is simulated by the number of stripes (0, 7, 11, 20) per image, i.e., from zero to severe occlusion.

{\noindent \bf Rotation.} To present realistic object appearance change, the object rotates in 3D in addition to position change. 
The rotation dynamics is specified by the angular velocity along each axis, which is kept constant throughout the sequence. 
We identify four rotation speed levels, (0, 1.3, 5.4, 10.6) degrees per frame, thus ranging from no rotation to fast rotation.  

\subsection{Trans2k dataset generation} \label{sec:range_identification}

Our preliminary study~\citep{trans2k_bmvc2022} reveals that most of the attribute levels described in Section~\ref{sec:attributes} result in performance reduction and are thus kept as relevant in our final dataset.
Two attribute levels including the lowest transparency level and zero rotation were identified as already well addressed by the opaque object training sets and are thus omitted from the dataset for better use of its capacity.

The following parameters are thus applied when rendering Trans2k. The GoT10k~\citep{got10k} training set sequences are sampled at random and at most once. All object types are sampled with equal probability. 
The transparency levels (excluding the lowest level) are sampled with equal probability. Blur presence in a sequence is sampled with 0.15 probability, with blur levels sampled uniformly. 
Occlusion presence is sampled with 0.2 probability, while occlusion levels are sampled uniformly. Rotation level is uniformly sampled. 
The resulting training dataset Trans2k thus contains 2,039 challenging sequences and 104,343 frames in total.

\begin{figure}[t]
\centering
\includegraphics[width=\linewidth]{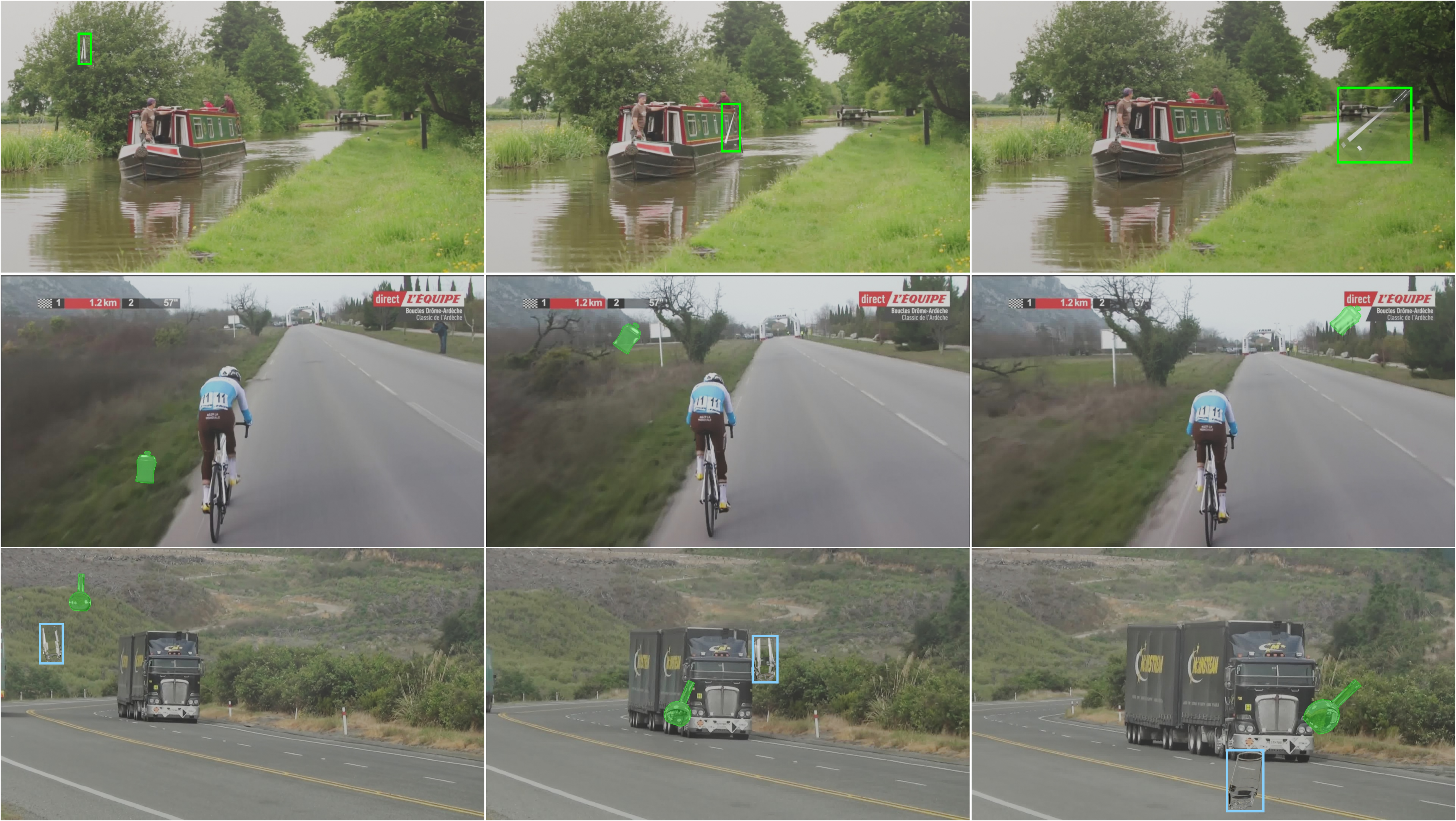}
\caption{Targets in Trans2k are annotated by axis-aligned bounding boxes (first row) or by segmentation masks (second and third rows). The dataset also contains annotated distractors (third row).
}
\label{fig:trans2k_qualitative}
\end{figure}

Since the sequences are rendered, the ground truth can be exactly computed. 
We provide the ground truth in two standard forms, the widely accepted target enclosing axis-aligned bounding-box and the segmentation mask to cater to the emerging segmentation trackers, e.g., see~\citep{kristan_vot2020}. 
The ground truths for distractors are generated as well.  
Trans2k is thus the first dataset with per-frame distractor annotation to facilitate development of distractor-aware methods. 
Some qualitative examples of the generated Trans2k sequences are shown in Figure~\ref{fig:trans2k_qualitative}. 

%%==================================%%
%% DiTra %%
%%==================================%%
\section{A distractor-aware tracker}  \label{sec:DiTra}

\begin{figure*}
  \centering
  \includegraphics[width=\linewidth]{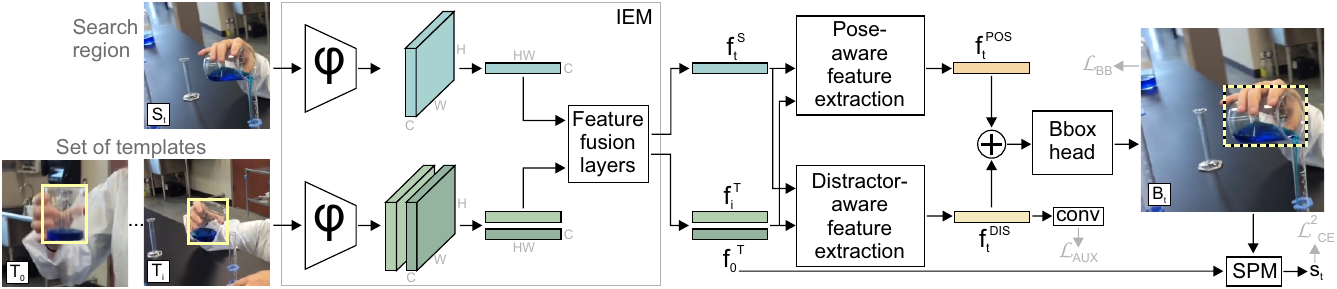}
  \caption{Overview of the proposed DiTra architecture. Features are first extracted from the search region and from a set of templates by Image encoding module (IEM). These features are then processed by two parallel branches generating pose-aware and distractor-aware features ($\mathbf{f}_{t}^{POS}$ and $\mathbf{f}_{t}^{DIS}$). Both features are summed together and processed by a bounding box prediction head to predict the target bounding box $\mathbf{B}_t$. Localization confidence score $s_t$ is estimated using the score prediction module (SPM).}
  \label{fig:overall-architecture}
\end{figure*}

We now introduce our second contribution -- a distractor-aware transparent object tracker DiTra (Figure~\ref{fig:overall-architecture}). 
Given $N_\mathrm{T}$ target templates $\mathbf{T} \in \{ \mathbf{T}_i \}_{i=1:N_\mathrm{T}}$ and their bounding boxes $\mathbf{B} \in \{ \mathbf{B}_i \}_{i=1:N_\mathrm{T}}$, DiTra localizes the target in the search region $\mathbf{S}_{t}$ at the current time-step $t$ by predicting a target bounding box $\mathbf{B}_t$.

The templates and the search region are of the same spatial size  (i.e., $H_{im} \times W_{im} \times 3$) and are first encoded by the Image encoding module (Section~\ref{sec:fem}). 
Next, the search region features are transformed into two sets of features by separate computational branches. 
The distractor-aware branch (Section~\ref{sec:distractor}) extracts features specialized only for discriminating between the target and similar objects. 
In parallel the pose-aware branch (Section~\ref{sec:pose}) extracts features tuned for maximally accurate pose estimation.

The two types of extracted features are then fused using a Target localization head (Section~\ref{sec:localization}) and regressed into the final estimated bounding box. 
Finally, a target presence confidence score is computed and the target template set is updated (Section~\ref{sec:template-update}). 
The following subsections detail each of the aforementioned computational blocks.

\subsection{Image encoding module}  \label{sec:fem}

The Image encoding module, IEM, (Figure~\ref{fig:overall-architecture}) first encodes the RGB templates and the search region by passing each through a backbone network $\varphi(\cdot)$, e.g., ResNet~\citep{resnet_cvpr16} and reduces the dimensionality of the output by a $1 \times 1$ convolution  to $C$ channels. 
This is followed by $L$ transformer-based {\it Feature fusion layers}, adopted from~\cite{transt_cvpr2021}. 
The templates and the search region are thus mapped into template and search region features, i.e.,  $\mathbf{T} \rightarrow \mathbf{F}^T \in \{ \mathbf{f}_{i}^{T} \}_{i=1:N_\mathrm{T}}$ and $\mathbf{S}_t \rightarrow \mathbf{f}_{t}^{S}$, each the size of $HW \times C$. 
The search region is then further encoded separately by two feature extraction branches described in the following two subsections.

\subsection{Distractor-aware feature extraction} \label{sec:distractor}

The task of this branch is to extract features focusing on discriminating the target from similar objects in its vicinity. 
We exploit the fact that the distractors will appear in the larger neighborhood of the localized targets in the previous frames. 
Assuming frequent template updating, distractors can be captured by constructing sufficiently large templates. 
In our tracker we thus set the template size equal to the search region. 

The distractor-aware feature extraction branch is implemented as a single multi-head attention block~\citep{attention_nips2017}. 
The search region features $\mathbf{f}_{t}^{S}$ are used as queries, the template features $\mathbf{f}_{i}^{T}$ as keys, while the values are obtained by summing the template features and the template encodings $\mathbf{f}_{i}^{E}$. 
The latter enables the attention mechanism to distinguish the target from the potential distractors.
The output of the multi-head attention block is followed by two linear transformations and ReLu activations to produce the final distractor-aware features $\mathbf{f}_{t}^{DIS} \in \mathbb{R}^{HW \times C}$. 

The template encodings are computed as follows. For each template a two-channel binary mask $\hat{\mathbf{m}}_i \in \mathbb{R}^{H W \times 2}$ is constructed. 
The values in the first channel are set to one within the target bonding box and to zero elsewhere, while the second channel is the inverse of the first. 
The mask is then linearly transformed into the template encoding $\mathbf{f}_i^{E}$.

\subsection{Pose-aware feature extraction}  \label{sec:pose}

The features specializing on discriminating the target from the distractors are extracted by the distractor-aware branch. 
This allows exploiting the entire capacity of the pose-aware branch solely for bounding box estimation, without compromising discrimination between similar objects, handled already by the distractor-aware branch.

To facilitate bounding box prediction learning, the templates are cropped to contain only the target appearance and not the potential distractors in their vicinity. 
Cropping thus converts the backbone template features  $\mathbf{f}_{i}^{T}$ into $\mathbf{f}_{i}^{*T} \in \mathbb{R}^{hw \times C}$. 
The features are then processed by a single multi-head attention block~\citep{attention_nips2017}, where the search region $\mathbf{f}_{t}^{S}$ is used as a query and templates $\mathbf{f}_{i}^{*T}$ are used as keys and values. 
This is followed by two linear transformations and ReLu activation to produce the final target pose-aware features $\mathbf{f}_{t}^{POS} \in \mathbb{R}^{HW \times C}$. 

\subsection{Target localization}  \label{sec:localization}

The target location is predicted by considering both positional and discriminative features, which are summed into localization features, i.e., $\mathbf{f}_{t}^{LOC} = \mathbf{f}_{t}^{POS} + \mathbf{f}_{t}^{DIS}$,
and a convolutional bounding box head~\citep{stark_iccv21} is applied to 
predict a single bounding box $\mathbf{B}_t \in \{ x_t, y_t, w_t, h_t \}$, where $x_t, y_t$ are coordinates of the top-left corner and $w_t, h_t$ are width and height, respectively. 

\subsection{Updating the template set}  \label{sec:template-update}

Target localization requires multiple templates to represent the target appearance, which can significantly change during tracking. 
Thus the set of templates $\mathbf{T}$ is dynamically updated by adding a new template in the set every 10 frames. 
When the number of the templates exceeds $N_\mathrm{T}$, the oldest one is removed, 
while keeping the initial template extracted in the first frame always within the set.

Since the distractors that lead to tracking failures are likely positioned in the vicinity of the target in the previous frame, the set is additionally updated by a 
template extracted at the previous frame. This template, however, is only used in the distractor-aware feature computation and not in the pose-aware feature extraction.

Tracking quality highly depends on ensuring targets are well localized in the templates.
Updating the template set when the target is poorly localized, (i.e., during occlusion, momentary drift or target absence) can lead to tracking failure. 
DiTra thus estimates the confidence score $s_t \in (0, 1)$ after the target localization step and updates the template set only if the confidence is high enough, i.e., $s_t > 0.5$. 

The confidence score $s_t$ is estimated by the score prediction module (SPM)~\citep{mixformer_cvpr2022} as follows. 
The localized target appearance is encoded by a learnable token attended to the search region features $\mathbf{f}_{t}^{S}$ extracted from the estimated bounding box $\mathbf{B}_t$. 
Next, the token is attended to the features from the initial target template $\mathbf{f}_0^{*T}$ and regressed into $s_t$ by a MLP with a sigmoid function. 
The reader is referred to~\cite{mixformer_cvpr2022} for additional details. 

\subsection{Training details}  \label{sec:training}
 
DiTra is trained in two phases.
The first phase is dedicated to training robust and accurate target localization, while the second phase is dedicated to training the target presence prediction module SPM (Section~\ref{sec:template-update}).

\noindent\textbf{Phase 1}. 
The whole network (except the SPM) is trained for target localization by optimizing the following localization loss: 
\begin{equation}  \label{eq:bbox_loss}
    \mathcal{L}_{BB} = \lambda_{GIOU} \mathcal{L}_{GIOU}(\mathbf{B}_t, \mathbf{B}_{GT}) + \lambda_{L1}\mathcal{L}_{1}(\mathbf{B}_t, \mathbf{B}_{GT}), 
\end{equation}
where $\mathbf{B}_t$ and $\mathbf{B}_{GT}$ are predicted and ground-truth bounding boxes, respectively, $\mathcal{L}_{GIOU}$ represents generalized IoU loss~\citep{giou_cvpr2019} and $\mathcal{L}_{1}$ is the $\ell_1$ loss.
The losses are weighted by $\lambda_{GIOU} = 2$ and $\lambda_{L1} = 5$. 

To guide the network towards learning distractor-aware features, we 
add an auxiliary loss $\mathcal{L}_{AUX}$ to the output of the distractor-aware feature extraction block. 
A $1 \times 1$ convolution, denoted as $\phi(\cdot)$ is first used to map the distractor-aware features $\mathbf{f}_{t}^{DIS}$ to a single channel segmentation mask. 
The auxiliary loss is then computed as a standard cross-entropy loss $\mathcal{L}_{CE}(\cdot)$, i.e.,
\begin{equation}  \label{eq:aux}
    \mathcal{L}_{AUX} = \mathcal{L}_{CE}(\phi(\mathbf{f}_{t}^{DIS}), \mathbf{m}_{t}^{GT}),
\end{equation}
where $\mathbf{m}_{t}^{GT}$ is obtained by setting pixels within the ground truth bounding box to one and zero outside.
The first-stage final training loss is then composed of the individual losses, i.e., $\mathcal{L} = \mathcal{L}_{BB} + \mathcal{L}_{AUX}$.
Combination of the two losses $\mathcal{L}_{BB}$ and $\mathcal{L}_{AUX}$ guides the network pose- and distractor-aware feature extraction branches to focus on their individual tasks.

\noindent\textbf{Phase 2}.
Only the score prediction module (SPM) is trained in this phase by minimizing the cross-entropy loss
\begin{equation}  \label{eq:cross-entropy}
    \mathcal{L}^{2}_{CE} = y_t log(s_t) + (1 - y_t) log(1 - s_t),
\end{equation}
where $s_t$ is the predicted target presence confidence score and $y_t \in \{ 0, 1 \}$ is the ground-truth label of the $t$-th training sample, i.e., whether the search region contains the target or not. 
Note that the superscript 2 in $\mathcal{L}^{2}_{CE}$ denotes training in the second phase.

%%==================================%%
%% Experiments: Trans2k %%
%%==================================%%
\section{Validation of Trans2k}  \label{sec:experiments_trans2k}

This section reports empiric validation of the proposed Trans2k training dataset.
A set of trackers (Section~\ref{sec:trackers}) is trained on Trans2k and evaluated against their versions trained on opaque object training sets (Section~\ref{sec:baseline_comparison} ).

\subsection{Trackers and training setup}  \label{sec:trackers}

State-of-the-art learning-based trackers that cover the major trends in modern architecture designs are selected:
(i) two siamese trackers SiamRPN++~\citep{siamrpn_cvpr2019} and SiamBAN~\citep{siamban_cvpr20}, 
(ii) two deep correlation filter trackers ATOM~\citep{atom_cvpr19} and DiMP~\citep{danelljan_dimp_iccv19}, 
(iii) the recent state-of-the-art transparent object tracker TransATOM~\citep{totb_iccv21}, and 
(iv) a transfomer-based STARK~\citep{stark_iccv21}. 
These trackers localize the target by a bounding box. To account for the recent trend in localization by per-pixel segmentation~\citep{kristan_vot2020}, we include 
(v) the recent state-of-the-art segmentation-based tracker D3S~\citep{Lukezic_CVPR_2020}.

For training on Trans2k, the trackers were initialized by the pre-trained weights provided by their authors, while all the training details were the same as in the original implementations. 
The trackers were trained for 50 epochs with 10000 training samples per epoch. Since Trans2k was designed as a complementary dataset covering situations not present in existing datasets, the training considers samples from Trans2k as well as opaque object sequences. 
In particular, we merged the opaque object training datasets GoT10k~\citep{got10k}, LaSoT~\citep{lasot_cvpr19} and TrackingNet~\citep{muller_trackingnet} into a single dataset, abbreviated as an \textit{opaque object training dataset} (OTD). 
A training batch is then constructed by sampling from Trans2k and OTD with 5:3 ratio.

\subsection{Results} \label{sec:baseline_comparison}

\begin{figure*}[t]
\centering
\includegraphics[width=\linewidth]{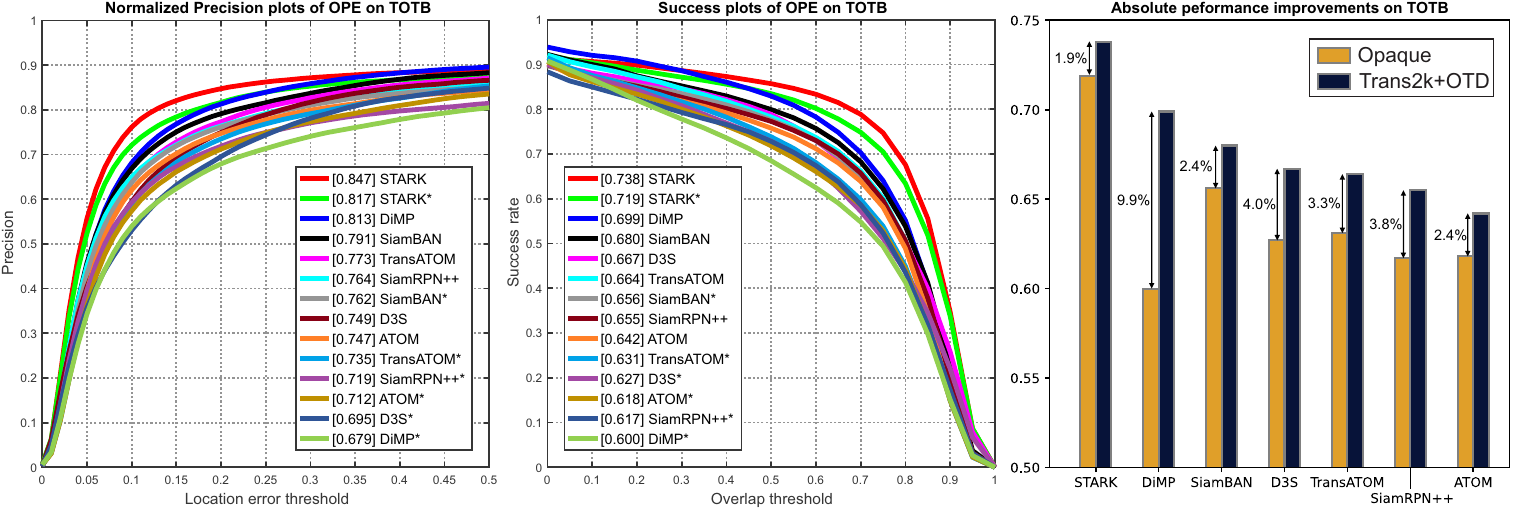}
\caption{Trackers evaluated on TOTB dataset shown in precision and success plots. Trackers trained on opaque datasets only are denoted by a star (*). The right graph shows absolute improvements in tracking performance measured by the AUC measure after training with the proposed Trans2k.}
\label{fig:baseline_comparison}
\end{figure*}

The contribution of the new training dataset Trans2k is validated by measuring tracking performance on the recent transparent object tracking benchmark TOTB~\citep{totb_iccv21}. 
Following the training regime described in Section~\ref{sec:trackers} the selected trackers were re-trained using Trans2k. 
Their performance was then compared to their original performance, i.e., when trained only with opaque object tracking sequences. 
Thus any change in performance is contributed only by the training dataset.
The trackers were evaluated by the standard one-pass evaluation protocol (OPE) that quantifies the performance by the AUC and center error measures on success and precision plots. 
For more information on the evaluation protocol, please refer to~\cite{otb_pami2015} and~\cite{totb_iccv21}. 

The results are shown in Figure~\ref{fig:baseline_comparison}. The performance of all trackers substantially improved when trained using Trans2k. The performance gains are at a level usually expected for a clear methodological improvement. 
Recently, TransATOM~\citep{totb_iccv21}, a transparent object tracking extension of ATOM~\citep{atom_cvpr19}, was proposed, which outperformed ATOM by 2.1\%. 
Without any methodological modification and only training with Trans2k, ATOM \textit{outperforms} this extension by 1.7\%. Nevertheless, TransATOM gains 3.3\% when trained with Trans2k.
The largest performance boost is achieved by DiMP, which improves by over 16\% and scores as the second-best among all the tested trackers.

Since Trans2k provides segmentation ground truths in addition to bounding boxes, it boosts the segmentation-based tracker D3S~\citep{Lukezic_CVPR_2020} as well. 
The version trained with Trans2k gains a remarkable 6\% in performance. 

Consistent with the observation on opaque object tracking benchmarks, the transformer-based tracker STARK achieves the best performance among existing trackers. 
Note that even without training with Trans2k, these tracker surpasses all (non-transformer) trackers. 
When trained with Trans2k, additional performance boost of 2.5\% is observed. 

%%==================================%%
%% Experiments: DiTra %%
%%==================================%%
\section{Validation of DiTra}  \label{sec:experiments_ditra}

This section provides an experimental evaluation of the proposed distractor-aware transparent object tracker (DiTra). 
Implementation details are given in Section~\ref{sec:implementation_details}, while DiTra is evaluated on transparent and opaque object tracking in Section~\ref{sec:experiments_transparent} and Section~\ref{sec:experiments_opaque}, respectively. 
Ablation study is reported in Section~\ref{sec:ablation}.

\subsection{Implementation details}  \label{sec:implementation_details}

\noindent\textbf{Tracking implementation details}.
Features extracted from the fourth layer of the ResNet-50~\citep{resnet_cvpr16} pre-trained on ImageNet~\citep{imagenet_ijcv_2015} for object classification are used in DiTra Image encoding module. 
The backbone features are extracted from the image region resized to $H_{im} = W_{im} = 320$ pixels, while the spatial dimensions of the features are reduced 16-times, i.e., $H = W = 20$. 
The channel dimension $C$ is set to 256. 

$N_\mathrm{T}=6$ templates are used in tracking. 
All attention blocks in DiTra contain 8 heads and the standard sine-based positional embeddings~\citep{attention_nips2017, detr_eccv2020, stark_iccv21} are used on queries and keys.
Tracking performance of the proposed tracker is not sensitive to the exact values of the parameters, thus we use the same values in all experiments.
\newline

\noindent\textbf{Training implementation details}.
As described in Section~\ref{sec:training}, the training process is divided into two phases.
In the {\it phase 1}, a search region is randomly sampled from a random training sequence and two templates are sampled within 200 frames from the same sequence. 
In the {\it phase 2} (i.e., training the score prediction module -- SPM), the positive and negative training samples are sampled with equal probability.
A positive sample is constructed by sampling a template and search region from the same sequence, while the negative sample is constructed by sampling them from different sequences. 

In the phase 1, DiTra is trained for 300 epochs using ADAM optimizer~\citep{adam_2015} with learning rate set to $10^{-4}$ decreasing by factor 10 after 250 epochs. 
Training takes approximately 4 days on two NVidia V100 with batch size 32 per-gpu.
In phase 2, DiTra is trained for 40 epochs using ADAM optimizer~\citep{adam_2015} with learning rate set to $10^{-4}$ decreasing by factor 10 after 30 epochs. 
Training takes approximately 8 hours on two NVidia V100 with batch size 64 per-gpu.

\subsection{Transparent object tracking}  \label{sec:experiments_transparent}

Transparent object tracking performance is evaluated on the recent TOTB benchmark~\citep{totb_iccv21}. 
The following state-of-the-art trackers are considered for comparison: 
two siamese trackers SiamRPN++~\citep{siamrpn_cvpr2019} and SiamBAN~\citep{siamban_cvpr20}, 
three deep correlation filter trackers ATOM~\citep{atom_cvpr19}, DiMP~\citep{danelljan_dimp_iccv19} and KYS~\citep{kys_eccv2020}, 
the recent state-of-the-art transparent object tracker TransATOM~\citep{totb_iccv21}, two transfomer-based trackers STARK~\citep{stark_iccv21} and TOMP~\citep{tomp_cvpr2022} and a segmentation-based tracker D3S~\citep{Lukezic_CVPR_2020}. 
The trackers are evaluated by the standard one-pass evaluation protocol (OPE) that quantifies the performance by the AUC score~\citep{totb_iccv21}. 

\begin{figure}[t]
  \centering
  \includegraphics[width=\linewidth]{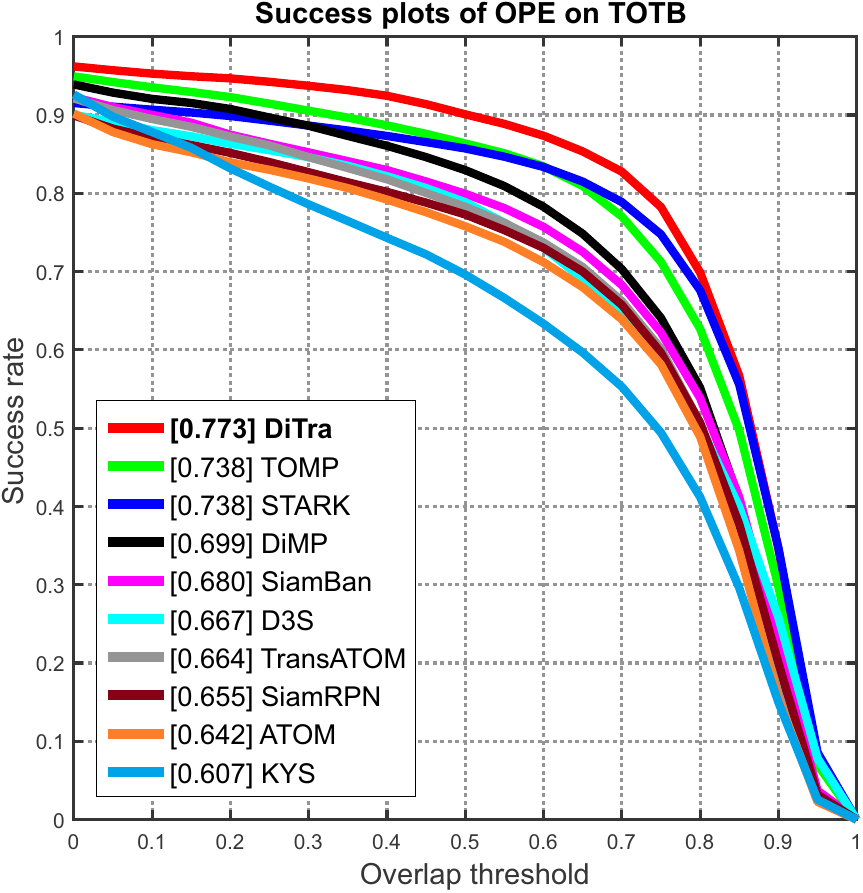}
  \caption{Performance on transparent object tracking benchmark TOTB~\citep{totb_iccv21}.}
  \label{fig:totb_transparent}
\end{figure}

\begin{figure}[h]
  \centering
  \includegraphics[width=\linewidth]{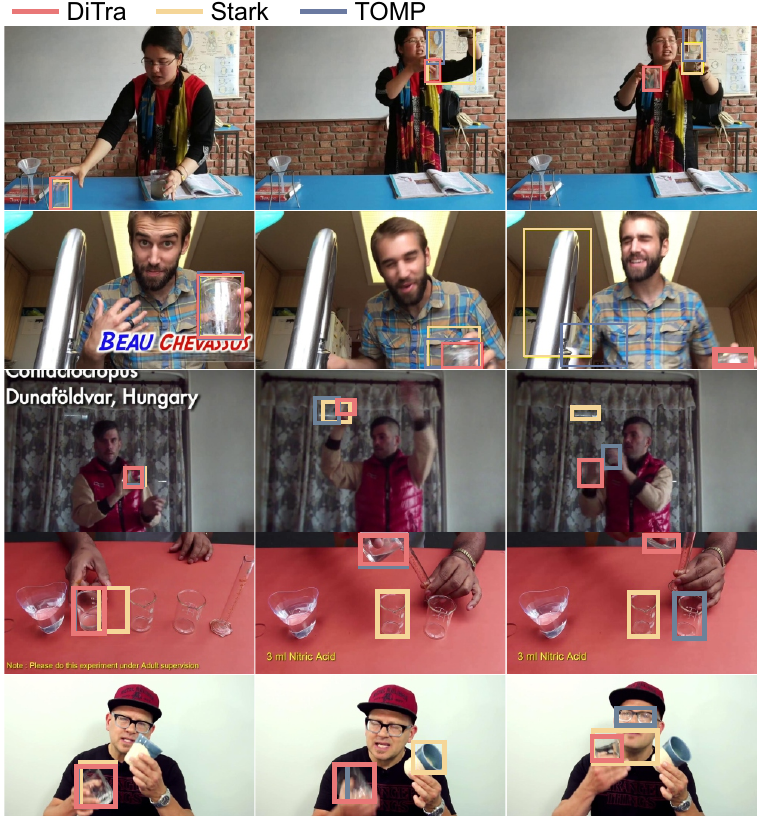}
  \caption{Qualitative comparison of DiTra, Stark and TOMP on TOTB~\citep{totb_iccv21}.}
  \label{fig:qualitative_sota}
\end{figure}

Results reported in Figure~\ref{fig:totb_transparent} show that the proposed DiTra outperforms all trackers and sets new state-of-the-art on TOTB. 
In particular, it outperforms the second-best STARK and TOMP for approximately $5\%$ in AUC. 
These results show that the disctractor-aware mechanism in the proposed DiTra successfully handles distractors in challenging scenarios and represents a strong baseline for the further research in tracking transparent objects. 
A qualitative comparison of DiTra with state-of-the-art trackers evaluated on TOTB is shown in Figure~\ref{fig:qualitative_sota}.


\subsection{Opaque object tracking}  \label{sec:experiments_opaque}

For evaluation completeness, DiTra is evaluated on opaque object tracking problems as well. 
The trackers are first evaluated on the challenging VOT2020 dataset~\citep{kristan_vot2020},
which is part of the annual VOT challenges~\citep{kristan_vot_tpami2016}. 
Trackers are run on each sequence multiple times from different pre-defined starting points and let to track until the end of the sequence. 
Tracking performance is measured by two complementary measures: (i) accuracy, computed as the average overlap and (ii) robustness, which counts how often tracker fails to localize the target. 
Both, accuracy and robustness are combined in the primary measure, called expected average overlap (EAO).

DiTra achieves top results among the compared trackers, in particular it outperforms the second-best STARK by $2\%$ EAO. 
Results show that DiTra particularly excells in robustness. This is due to the discriminative formulation, which allows to resolve challenging situations with multiple distractors.

Next, we evaluate DiTra on GoT10k~\citep{got10k} test dataset, which is a large-scale high-diversity tracking dataset. 
It consists of approximately 10 thousand training sequences, while a set of 180 sequences is used for evaluating tracking performance. 
A tracker is initialized at the beginning and let to track to the end of the sequence. 
Tracking performance is measured by the area under the success-rate curve (AUC). 
Results in Table~\ref{tab:comparison-got-lasot} show that DiTra outperforms the compared recent state-of-the-art tracker~\citep{mixformer_cvpr2022} by nearly 4$\%$. 
This result demonstrates that DiTra achieves state-of-the-art results on opaque object tracking and also generalizes well across different short-term datasets.

Despite being a short-term tracker, we evaluate DiTra on the long-term tracking dataset LaSoT~\citep{lasot_cvpr19}. 
In this dataset, the targets often disappear from the image, which emphasizes {\it long-term capabilities} of a tracker.
The dataset contains a total of 1400 sequences with 70 object categories, where 280 sequences are used for evaluation and others are used for training. 
A tracker is initialized at the beginning and let to track to the end of the sequence. 
Tracking performance is measured by the area under the success-rate curve (AUC). 
 
Results in Table~\ref{tab:comparison-got-lasot} show that DiTra performs comparable to the top-performing MixFormer~\cite{mixformer_cvpr2022}, TOMP~\citep{tomp_cvpr2022} and STARK~\citep{stark_iccv21}. 
Based on the results obtained on VOT, GoT10k and LaSoT datasets, we conclude that DiTra excels both in tracking of transparent objects as well as opaque objects, indicating the generality of the proposed distractor-aware formulation.

\begin{table}
  \centering
  \caption{Performance on the VOT2020~\cite{kristan_vot2020} opaque object tracking benchmark.}
  \label{tab:comparison-vot}
  \begin{tabular}{lccc}
    \toprule
    Tracker & EAO & Accuracy & Robustness \\
    \midrule
    {\bf DiTra} & 0.314 & 0.447 & 0.821 \\
    STARK & 0.308 & 0.478 & 0.799 \\
    TOMP & 0.297 & 0.453 & 0.789 \\
    DiMP & 0.274 & 0.457 & 0.734 \\
    ATOM & 0.271 & 0.462 & 0.734 \\
    SiamRPN++ & 0.255 & 0.424 & 0.730 \\
    \bottomrule
  \end{tabular}
\end{table}

\begin{table}
  \centering
  \caption{Performance on opaque tracking datasets GoT10k~\citep{got10k} and LaSoT~\citep{lasot_cvpr19}.
  }
  \label{tab:comparison-got-lasot}
  \begin{tabular}{lcc}
    \toprule
    Tracker & GoT10k & LaSoT \\
    \midrule
    {\bf DiTra} & 76.1 & 66.0 \\
    MixFormer-1k & 73.2 & 67.9 \\
    STARK & 68.0 & 66.4 \\
    TOMP & 67.0 & 67.6 \\
    KYS & 63.6 & 55.4 \\
    DiMP & 61.1 & 56.9 \\
    ATOM & 55.6 & 51.5 \\
    SiamRPN++ & 51.7 & 49.6 \\
    \bottomrule
  \end{tabular}
\end{table}

\subsection{Ablation study}  \label{sec:ablation}

Ablation study on the TOTB benchmark~\citep{totb_iccv21} is conducted for further insights.
The following variations of DiTra are analyzed:
(i)~DiTra without fine-tuning on transparent objects (DiTra$^{\overline{TRS}}$), i.e., trained on opaque objects only; 
(ii)~DiTra without the distractor-aware feature extraction branch (DiTra$^{\overline{DIS}}$); 
(iii)~DiTra without the pose-aware feature extraction branch (DiTra$^{\overline{POS}}$); and 
(iv)~DiTra without the most recent template in the distractor-aware feature extraction (DiTra$^{\overline{REC}}$). 
Note that the variants (ii) and (iii) are trained using the same training setup as the original DiTra, while the variant (iv) does not require re-training. 

\begin{figure}[t]
  \centering
  \includegraphics[width=\linewidth]{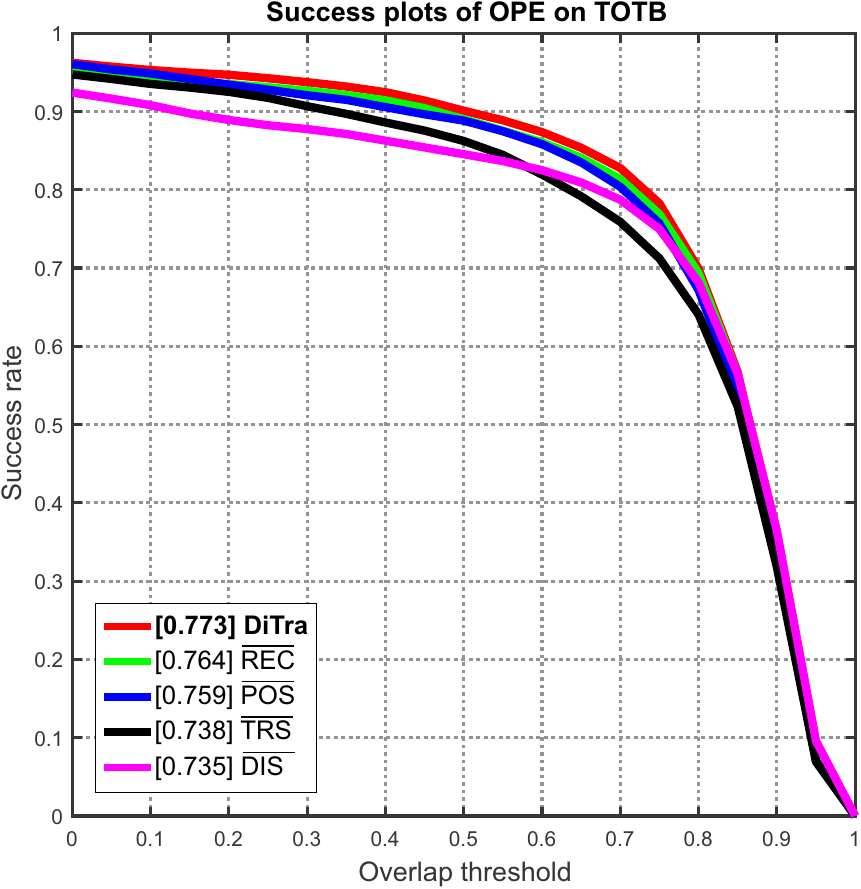}
  \caption{Ablation study on TOTB. Removing: the most recent template ($\overline{REC}$), the pose-aware feature ($\overline{POS}$) and the distractor-aware features ($\overline{DIS}$). The ($\overline{TRS}$) denotes a version of DiTra without fine-tuning on transparent objects.}
  \label{fig:ablation}
\end{figure}

Results of the ablation study are presented in Figure~\ref{fig:ablation}. 
Omitting the fine-tuning step on transparent objects (DiTra$^{\overline{TRS}}$) reduces the tracking performance by $4.5\%$. This confirms the contribution of the Trans2k dataset and shows the importance of including transparent objects in training the process. 

Removing the distractor-aware feature extraction branch (DiTra$^{\overline{DIS}}$) causes a $5\%$ performance drop, while 
removal of the pose-aware feature extraction branch (DiTra$^{\overline{POS}}$) reduces the performance by $2\%$. 
These results show the importance of splitting the tracking task into two separate branches. 
However, the distractor-aware features are more important for good tracking performance than the pose-aware features, since they prevent irreversible tracking failures.

Finally, removing the most recent template in the distractor-aware feature extraction (DiTra$^{\overline{REC}}$) reduces the tracking performance by approximately $1\%$. 
This demonstrates that the most recent template is not essential, but helps when target appearance is changing quickly in presence of multiple distractors. 

\begin{figure}[!t]
  \centering
  \includegraphics[width=\linewidth]{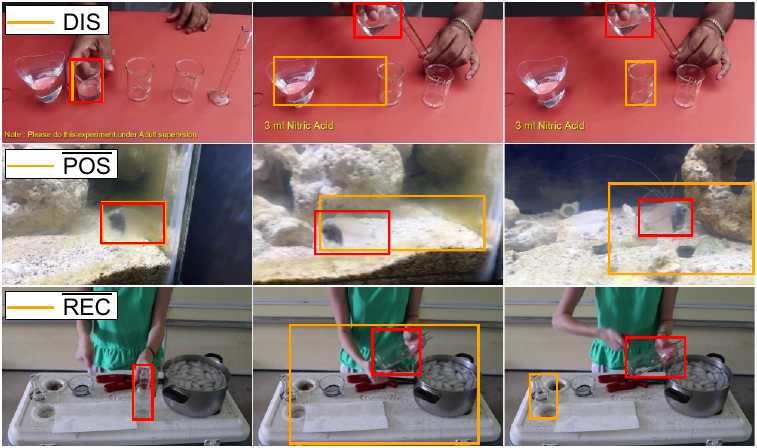}
  \caption{The proposed DiTra (red bounding box) is compared to the versions: without distractor-aware features ($\overline{DIS}$), without pose-aware features ($\overline{POS}$) and without the most recent template ($\overline{REC}$). Please see text in Section~\ref{sec:ablation} for discussion.}
  \label{fig:qualitative}
\end{figure}

Qualitative comparison is given in Figure~\ref{fig:qualitative}.
Removing the distractor-aware feature extraction branch (DiTra$^{\overline{DIS}}$) reduces the tracking capability especially when multiple similar objects (distractors) are present in the same scene (third row). 
A version without the pose-aware feature extraction branch (DiTra$^{\overline{POS}}$) fails to accurately localize the target in challenging scenarios (fourth row). 
Removal of the most recent template in the distractor-aware feature extraction branch (DiTra$^{\overline{REC}}$) causes tracking failure when the target appearance changes significantly in presence of distractors (fifth row). 

\begin{figure}[!t]
  \centering
  \includegraphics[width=\linewidth]{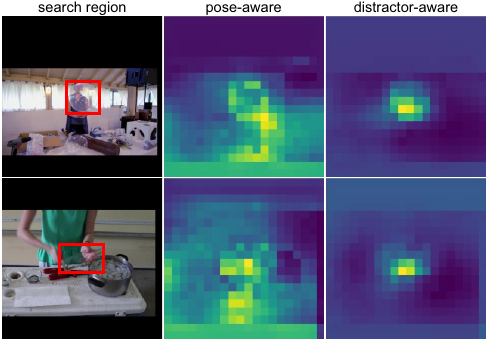}
  \caption{Visualization of the attention maps of the pose-aware and distractor-aware feature extraction for the corresponding search regions.}
  \label{fig:attention-maps}
\end{figure}

To provide additional insights of the proposed tracker, we visualize the pose-aware and distractor-aware feature extraction attention maps in Figure~\ref{fig:attention-maps}. 
Attention operation of the pose-aware feature extraction focuses on object shape, highlighting shapes of individual (multiple) objects, not only the target. 
On the other hand, attention of the distractor-aware feature extraction successfully suppresses distractors and only provides the object center. 
Combination of both, the pose-aware and distractor-aware feature extraction results in an accurate and robust tracker.

\subsection{Failure cases}  \label{sec:failure-cases}

\begin{figure}[!h]
  \centering
  \includegraphics[width=\linewidth]{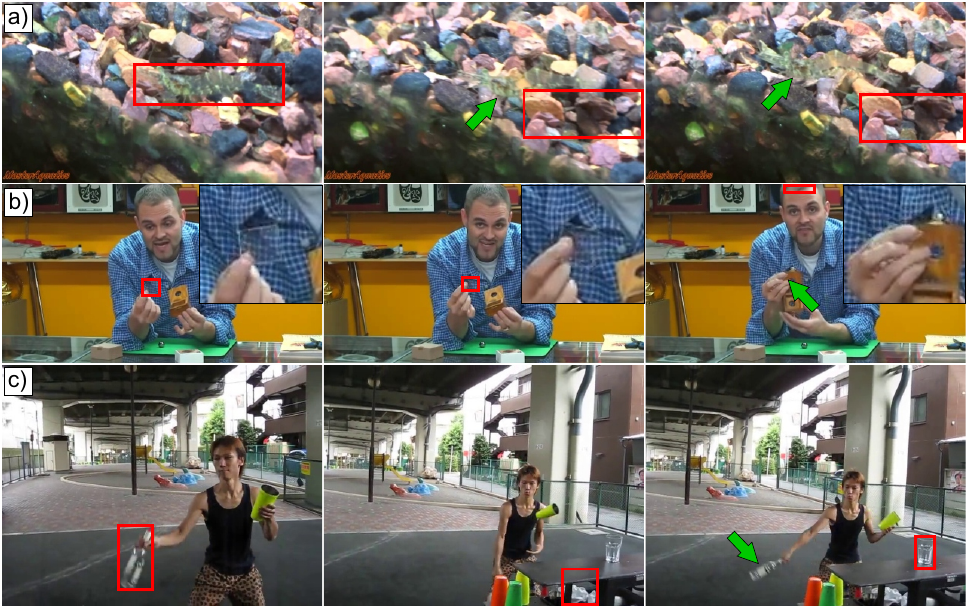}
  \caption{Failure cases of DiTra. Two most frequent reasons for a failure are extreme transparency of the target (a) and (b) and combination of occlusion and distractors (c).}
  \label{fig:failure-cases}
\end{figure}

An analysis of failure cases of the proposed tracker is presented in Figure~\ref{fig:failure-cases}. 
We observe two major reasons causing DiTra to fail. 
First one is an extreme level of transparency, which results in a poorly visible target. 
Two examples showing such objects are shown in Figure~\ref{fig:failure-cases}~(a) and~(b). 
In these examples DiTra tends to focus on the background, which is visible through the target, instead of tracking it. 
Note that such examples are extremely difficult for humans as well. 
We believe that these failures could be addressed by specializing the feature extractor and forcing it to focus on the really fine visual details, specific for such objects.

Another group of failures are situations where occlusion appears together with distractors, shown on Figure~\ref{fig:failure-cases}~(c). 
When the target gets occluded, the tracker localizes the object, which is visually the most similar to the target. 
If the target re-appears in the position outside of the search region, the tracker is not able to localize it and keeps tracking the wrong target (distractor). 
A possible solution to such failures would be to incorporate long-term tracking components, e.g., image-wide re-detection mechanism or motion priors. 

%%==================================%%
%% Conclusion %%
%%==================================%%
\section{Conclusion}  \label{sec:conclusion}

Two contributions to transparent object tracking were presented.
The first contribution is the first transparent object tracking training dataset Trans2k, which exploits 
the fact that transparent objects can be sufficiently realistically rendered by modern renderers. 
Trans2k was validated on the recent transparent object tracking benchmark TOTB~\citep{totb_iccv21}. Training with Trans2k improves performance at levels usually observed in fundamental methodological advancements in tracking algorithms. 
This behavior is observed over a wide range of tracking methodologies. 

The second contribution %of this paper 
is %the introduction of 
a new distractor-aware transparent object tracker (DiTra). 
DiTra addresses tracking in presence of multiple visually similar objects (distractors), which are common in transparent object tracking scenes. 
The proposed tracker achieves state-of-the-art performance on the transparent object tracking task and is competitive in opaque object tracking. Trans2k, its rendering engine and DiTra will be publicly released.

%While the field of transparent object tracking has recently obtained an excellent test set~\cite{totb_iccv21}, the main ingredient crucial for advancements, i.e., a curated training set was missing. 
While an excellent test set~\cite{totb_iccv21} was recently introduced for transparent object tracking, the second main ingredient crucial for advancements, i.e., a curated training set was missing. 
Trans2k fills this void and will enable future development of new learnable modules specifically addressing the challenges in transparent object tracking, thus fully unlocking the  power of modern deep learning trackers on this scientifically interesting domain. 
We envision that the Trans2k generation engine will allow innovative learning modes in which the sequences with specific challenges can be generated on demand to specialize the trackers to niche tasks or to improve their overall performance. In addition, the rendering 
engine could be used to generate training data for 6-DoF video pose estimation, thus benefiting research beyond 2D transparent object tracking.

Based on the excellent generalizaton to opaque object tracking, we hope that the proposed distractor-aware formulation in DiTra will ignite exploration of similar modules dedicated for opaque object tracking, thus leading to further advancements in both tracking sub-domains. 

\bibliography{egbib}

\end{document}